\documentclass[runningheads,a4paper]{llncs}

\usepackage{amssymb}
\usepackage{graphicx}
\usepackage{multirow}

\usepackage{url}
\urldef{\mailsa}\path|ehsansherkat@dal.ca, eem@cs.dal.ca|
\newcommand{\keywords}[1]{\par\addvspace\baselineskip
\noindent\keywordname\enspace\ignorespaces#1}

\begin{document}

\mainmatter  % start of an individual contribution

\title{Vector Embedding of Wikipedia Concepts and Entities}

% \titlerunning{Submitted to NLDB 2017}

\author{Ehsan Sherkat*%
% \thanks{Please note that the LNCS Editorial assumes that all authors have used
% the western naming convention, with given names preceding surnames. This determines
% the structure of the names in the running heads and the author index.}%
\and Evangelos Milios}
%
% \authorrunning{Submitted to NLDB 2017}

\institute{Faculty of Computer Science,\\
Dalhousie University, Halifax, Canada\\
\mailsa\\
% \mailsb\\
% \mailsc\\
% \url{}
}


% \toctitle{Lecture Notes in Computer Science}
% \tocauthor{Authors' Instructions}
\maketitle

\begin{abstract}
Using deep learning for different machine learning tasks such as image classification and word embedding has recently gained many attentions. Its appealing performance reported across specific Natural Language Processing (NLP) tasks in comparison with other approaches is the reason for its popularity. Word embedding is the task of mapping words or phrases to a low dimensional numerical vector. In this paper, we use deep learning to embed Wikipedia Concepts and Entities. The English version of Wikipedia contains more than five million pages, which suggest its capability to cover many English Entities, Phrases, and Concepts. Each Wikipedia page is considered as a concept. Some concepts correspond to entities, such as a person's name, an organization or a place. Contrary to word embedding, Wikipedia Concepts Embedding is not ambiguous, so there are different vectors for concepts with similar surface form but different mentions. We proposed several approaches and evaluated their performance based on Concept Analogy and Concept Similarity tasks. The results show that proposed approaches have the performance comparable and in some cases even higher than the state-of-the-art methods.
\end{abstract}

\section{Introduction}
Recently, many researchers \cite{mikolov2013distributed,pennington2014glove} showed the capabilities of deep learning for natural language processing tasks such as word embedding. Word embedding is the task of representing each term with a low-dimensional (typically less than 1000) numerical vector. Distributed representation of words showed better performance than traditional approaches for tasks such as word analogy \cite{mikolov2013distributed}. Some words are Entities, i.e. name of an organization, Person, Movie, etc.  On the other hand, some terms and phrases have a page or definition in a knowledge base such as Wikipedia, which are called Concepts. For example, there is a page in Wikipedia for “Data Mining” or “Computer Science” concepts. Both Concepts and Entities are valuable resources to get semantic and better making sense of a text. In this paper, we used deep learning to represent Wikipedia Concepts and Entities with numerical vectors. We make the following contributions:

\begin{itemize} 
\item Wide coverage of words and concepts: about 1.7 million Wikipedia concepts and near 2 million English words were embedded in this research, which is one of the highest number of concepts embedding currently exists, to the best of our knowledge. The Concept and words vectors are also publicly available for research purposes\footnote{\url{https://github.com/ehsansherkat/ConVec}}. We also used one of the latest versions of Wikipedia English dump to learn words embedding. Over time, each term may appear in different contexts, and as a result, it may have different embeddings so this is why we used one of the recent versions of the Wikipedia.

\begin{table}[]
\centering
\caption{Top similar terms to "amazon" based on Word2Vec and GloVe.}
\label{hqv:table2}
\resizebox{\textwidth}{!}{%
\begin{tabular}{l|l|l|l|l|l|l}
\hline
\textbf{Word2Vec} & itunes     & play.com   & cli       & adobe\_acrobat & amiga  & canada        \\ \hline
\textbf{GloVe}    & amazon.com & rainforest & amazonian & kindle         & jungle & deforestation \\ \hline
\end{tabular}%
}
\end{table}

\item Unambiguous word embedding: Existing word embedding approaches suffer from the problem of ambiguity. For example, top nine similar terms to 'Amazon' based on pre-trained Google's vectors in Word2Vec \cite{mikolov2013distributed} and GloVe \cite{pennington2014glove} models are in Table \ref{hqv:table2}. Word2Vec and GloVe are the two first pioneer approaches for word embedding. In a document, 'Amazon' may refer to the name of a jungle and not the name of a company. In the process of embedding, all different meaning of a word 'Amazon' is embedded in a single vector. Producing distinct embedding for each sense of the ambiguous terms could lead to better representation of documents. One way to achieve this, is using unambiguous resources such as Wikipedia and learning the embedding separately for each Entity and Concept. 

\item We compared the quality versus the size of the corpus on the quality of trained vectors. We demonstrated that much smaller corpora with more accurate textual content is better than a very large text corpora with less accuracy in the content for the concept and phrase embedding.

\item We studied the impact of fine tuning weights of network by pre-trained word vectors from a very large text corpora in tasks of Phrase Analogy and Phrase Similarity. Fine tuning is the task of initializing the weights of the network by pre-trained vectors instead of random initialization.

\item Proposing different approaches for Wikipedia Concept embedding and comparing results with the state-of-the-art methods on the standard datasets. 

\end{itemize}

% \begin{table}[]
% \centering
% \caption{Top similar terms to "amazon" based on Word2Vec and GloVe}
% \label{hqv:table2}
% \begin{tabular}{ll}
% \hline
% Word2Vec       & GloVe         \\ \hline
% itunes         & amazon.com    \\
% play.com       & rainforest    \\
% cli            & amazonian     \\
% adobe\_acrobat & kindle        \\
% amiga          & jungle        \\
% canada         & deforestation \\
% Amazon         & itunes        \\
% sony           & bezos         \\
% dvd            & brazil        \\ \hline
% \end{tabular}
% \end{table}

\section{Related Works}
Word2Vec and GloVe are two pioneer approaches for word embedding. Recently, other methods have been introduced that try to both improve the performance and quality of the word embedding \cite{faruqui2014improving} by using multilingual correlation. A method based on Word2Vec is proposed by Mikolov et al. for phrase embedding. \cite{mikolov2013distributed}. In the first step, they find the words that appear more frequently to gather  than separately, and then they replace them with a single token. Finally, the vector for phrases is learned in the same way as single word embedding. One of the features of this approach is that both words and phrases are in the same vector space.

Graph embedding methods \cite{cao2016deep} using Deep Neural Networks are similar to the goals of this paper. Graph representation has been used for information management in many real world problems. Extracting deep information from these graphs is important and challenging. One solution is using graph embedding methods. The word embedding methods use linear sequences of words to learn a word representation. For graph embedding, the first step is converting the graph structure to an extensive collection of linear sequences. A uniform sampling method named as Truncated Random Walk was presented in \cite{perozzi2014deepwalk}. In the second step, a word embedding method such as Word2Vec is used to learn the representation for each graph vertex. Wikipedia is also can be represented by a graph, and the links are the inter citation between Wikipedia's pages, called anchors.

A graph embedding method for Wikipedia using a similarity inspired by the HITS algorithm \cite{kleinberg1999authoritative} was presented by Sajadi et al. \cite{sajadi2015domain}. The output of this approach for each Wikipedia Concept is a fixed length list of similar Wikipedia pages and their similarity, which represents the dimension name of the corresponding Wikipedia concepts. The difference between this method and deep learning based methods is that each dimension of a concept embedding is meaningful and understandable by the human. 

A Wikipedia concept similarity index based on in-links and out-links of a page was proposed by Milne and Witten \cite{Milne2008}. In their similarity method, two Wikipedia pages are more similar to each other if they share more common in and out links. This method is used to compare the result of Concept similarity task with the proposed approaches.

The idea of using Anchor texts inside Wikipedia for learning phrase vectors is being used in some other researches \cite{tsai2016cross} as well. In this research, we proposed different methods to use anchor texts and evaluated the results in standard tasks. We also compared the performance of the proposed methods with top notch methods. 

\section{Distributed Representation of Concepts}
From this point on, we describe how we trained our word embedding. At first we describe the steps for preparing the Wikipedia dataset and then describe different methods we used to train words and Concepts vectors.

\paragraph{Preparing Wikipedia dataset: }  
In this research, the Wikipedia English Text used, is from Wikipedia dump May 01, 2016. In the first step, we developed a toolkit\footnote{\url{https://github.com/ehsansherkat/ConVec}} using several open source Python libraries (described in Appendix A) to extract all  pages in English Wikipedia, and as a result 16,527,332 pages were extracted. Not all of these pages are valuable, so we pruned the list by the several rules (Check Appendix B for more information). 

As a result of pruning, 5,001,168 unique Wikipedia pages, pointed at by the anchors, were extracted. For the next step, the plain text of all these pages were extracted in such a way that anchors belonging to the pruned list of Wikipedia pages were replaced (using developed toolkit) with their Wikipedia page ID (the redirects were also handled), and for other anchors, the surface form of them was substituted. We merged the plain text of all pages in a single text file in which, each line is a clean text of a Wikipedia page. This dataset contains 2.1B tokens.

\paragraph{ConVec: } 
The Wikipedia dataset obtained as a result of previous steps was used for training a Skip-gram model \cite{mikolov2013distributed} with negative sampling instead of hierarchical softmax. We called this approach as ConVec. The Skip-gram model is a type of Artificial Neural Network, which contains three layers:  input, projection, and output layer. Each word in the dataset is input to this network, and the output is predicting the surrounding words within a fixed window size (we used a window size of 10). Skip-gram has been shown to give a better result in comparison to the Bag of Words (CBOW) model \cite{mikolov2013distributed}. CBOW gets the surrounding words of a word and tries to predict the word (the reverse of the Skip-gram model). 

As a result of running the Skip-gram model on the Wikipedia dataset, we got 3,274,884 unique word embeddings, of which 1,707,205 are Wikipedia Concepts (Words and Anchors with a frequency of appearance in Wikipedia pages less than five are not considered). The procedure of training both words and Concepts in the same model result in Concepts and words belonging to the same vector space. This feature enables not only finding similar concepts to a concept but also finding similar words to that concept. 

\paragraph{ConVec Fine Tuned: } 
In Image datasets, it is customary to fine-tune the weights of a neural network with pre-trained vectors over a large dataset. Fine tuning is the task of initializing the weights of the network by pre-trained vectors instead of random initialization. We tried to investigate the impact of fine tuning the weights for textual datasets as well. In this case, we tried to fine-tune the vectors with Glove 6B dataset trained on Wikipedia and Gigaword datasets \cite{pennington2014glove}. The weights of the the skip-gram model initialized with GLove 6B pre-trained word vectors and then the training continued with the Wikipedia dataset prepared in the previous step. We called the Concept vectors trained based on this method ConVec Fine Tuned.

\paragraph{ConVec Heuristic: } 
We hypothesize that the quality of Concept vectors can improve with the size of training data. The sample data is the anchor text inside each Wikipedia page. Based on this assumption, we experimented with a heuristic to increase the number of anchor texts in each Wikipedia page. It is a Wikipedia policy that there is no self-link (anchor) in a page. It means that no page links to itself. On the other hand, it is common that the title of the page is repeated inside the page. The heuristic is to convert all exact mentions of the title of a Wikipedia page to anchor text with a link to that page. By using this heuristic, 18,301,475 new anchors were added to the Wikipedia dataset. This method called ConVec Heuristic.

\paragraph{ConVec Only Anchors: } 
The other experiment is to study the importance and role of the not anchored words in Wikipedia pages in improving the quality of phrase embeddings. In that case, all the text in a page, except anchor texts were removed and then the same skip-gram model with negative sampling and the window size of 10 is used to learn phrase embeddings. This approach (ConVec Only Anchors) is similar to ConVec except that the corpus only contains anchor texts. 

\paragraph{} An approach called Doc2Vec was introduced by Mikolov et al.\cite{le2014distributed} for Document embedding. In this embedding, the vector representation is for the entire document instead of a single term or a phrase. Based on the vector embeddings of two documents, one can check their similarity by comparing their vector similarity (e.g. Using Cosine distance). We tried to embed a whole Wikipedia page (concept) with its content using Doc2Vec and then consider the resulting vector as the Concept vector. The results of this experiment were far worse than the other approaches so we decided not to compare it with other methods. The reason is mostly related to the length of Wikipedia pages. As the size of a document increases the Doc2Vec approach for document embedding results in a lower performance.

\section{Evaluation}
Phrase Analogy and Phrase Similarity tasks are used to evaluate the different embedding of Wikipedia Concepts. In the following, detail results of this comparison are provided.

\paragraph{Phrase Analogy Task:}
To evaluate the quality of the Concept vectors, we used the phrase analogy dataset in \cite{mikolov2013distributed} which contains 3,218 questions. The Phrase analogy task involves questions like "\textit{Word1} is to \textit{Word2} as \textit{Word3} is to \textit{Word4}". The last word (Word4) is the missing word. Each approach is allowed to suggest the one and only one word for the missing word (Word4). The accuracy is calculated based upon the number of correct answers. In word embedding, the answer is finding the closest word vector to the Eq. \ref{eq1}. $V$ is the vector representation of the corresponding Word.

\begin{equation}
V_{Word2}  -  V_{Word1} +  V_{Word3} =  V_{Word4} \label{eq1}
\end{equation}

$V$ is the vector representation of the corresponding Word. The cosine is similarity used for majoring the similarity between vectors in each side of the above equation.

\begin{table}[]
\centering
\caption{Comparing the results of three different versions of ConVec (trained on Wikipedia 2.1B tokens) with Google Freebase pre-trained vectors over Google 100B tokens news dataset in the Phrase Analogy task. The Accuracy (All), shows the coverage and performance of each approach for answering questions. The accuracy for common questions (Accuracy (Commons)), is for fair comparison of each approach. \#phrases shows the number of top frequent words of each approach that are used to calculate the accuracy. \#found is the number of questions that all 4 words of them are present in the approach dictionary.}
\label{hqv:table3}
\begin{tabular}{l|l|cc|cc}
\hline
\multirow{2}{*}{Embedding Name} & \multicolumn{1}{c|}{\multirow{2}{*}{\#phrases}} & \multicolumn{2}{l|}{Accuracy (All)} & \multicolumn{2}{l}{Accuracy (Commons)} \\
 & \multicolumn{1}{c|}{} & \#found & Accuracy & \#found & Accuracy \\ \hline
\multirow{3}{*}{Google Freebase} & Top 30,000 & 1048 & 55.7\% & 89 & 52.8\% \\
 & Top 300,000 & 1536 & 47.0\% & 800 & 48.5\% \\
 & Top 3,000,000 & 1838 & 42.1\% & 1203 & 42.7\% \\ \hline
\multirow{3}{*}{ConVec} & Top 30,000 & 202 & 81.7\% & 89 & 82.0\% \\
 & Top 300,000 & 1702 & 68.0\% & 800 & 72.1\% \\
 & Top 3,000,000 & 2238 & 56.4\% & 1203 & 61.1\% \\ \hline
\multirow{3}{*}{\begin{tabular}[c]{@{}l@{}}ConVec \\ (Fine Tuned)\end{tabular}} & Top 30,000 & 202 & 80.7\% & 89 & 79.8\% \\
 & Top 300,000 & 1702 & 68.3\% & 800 & 73.0\% \\
 & Top 3,000,000 & 2238 & 56.8\% & 1203 & 63.6\% \\ \hline
\multirow{3}{*}{\begin{tabular}[c]{@{}l@{}}ConVec\\ (Heuristic)\end{tabular}} & Top 30,000 & 242 & 81.4\% & 89 & 80.9\% \\
 & Top 300,000 & 1804 & 65.6\% & 800 & 68.9\% \\
 & Top 3,000,000 & 2960 & 46.6\% & 1203 & 58.7\% \\ \hline
\end{tabular}
\end{table}

In order to calculate the accuracy in the Phrase Analogy, all four words of a question should be present in the dataset. If a word is missing from a question, the question is not included in the accuracy calculation. Based on this assumption, the accuracy is calculated using the Eq. \ref{eq2}.

\begin{equation}
Accuracy =  \frac{\#Correct Answers}{\#Questions With Phrases Inside Approach Vectors List} \label{eq2}
\end{equation}

We compared the quality of three different versions of ConVec with Google Freebase dataset pre-trained over Google 100B token news dataset. The skip-gram model with negative sampling is used to train the vectors in Google Freebase. The vectors in this dataset have 1000 dimensions in length. For preparing the embedding for phrases, they used a statistical approach to find words that appear more together than separately and then considered them as a single token. In the next step, they replaced these tokens with their corresponding freebase ID. Freebase is a knowledge base containing millions of entities and concepts, mostly extracted from Wikipedia pages.  

\begin{table}[]
\centering
\caption{Comparing the results in Phrase Similarity dataset. Rho is Spearman's correlation to the human evaluators. !Found is the number of pairs not found in each approach dataset.}
\label{hqv:table4}
\resizebox{1\textwidth}{!}{%
\begin{tabular}{lll|ll|ll|ll|ll}
\hline
\multicolumn{3}{c}{Datasets} & \multicolumn{2}{|l|}{\begin{tabular}[c]{@{}l@{}}Wikipedia\\ Miner \end{tabular}} & \multicolumn{2}{l|}{\begin{tabular}[c]{@{}l@{}}Google\\ Freebase\end{tabular}} & \multicolumn{2}{l}{ConVec} & \multicolumn{2}{|l}{\begin{tabular}[c]{@{}l@{}}ConVec\\ (Heuristic)\end{tabular}}\\ \hline
\# & Dataset Name & \begin{tabular}[c]{@{}l@{}}\#Pairs\end{tabular} & \begin{tabular}[c]{@{}l@{}}!Found\end{tabular} & Rho & \begin{tabular}[c]{@{}l@{}}!Found\end{tabular} & Rho & \begin{tabular}[c]{@{}l@{}}!Found\end{tabular} & Rho & \begin{tabular}[c]{@{}l@{}}!Found\end{tabular} & Rho\\ \hline
1 & WS-REL \cite{Finkelstein:2001} & 251 & 114 & 0.6564 & 87 & 0.3227 & 104 & 0.5594 & 57 & 0.5566 \\
2 & SIMLEX \cite{hill2016simlex} & 961 & 513 & 0.2166 & 369 & 0.1159 & 504 & 0.3406 & 357 & 0.2152 \\
3 & WS-SIM \cite{Finkelstein:2001} & 200 & 83 & 0.7505 & 58 & 0.4646 & 81 & 0.7524 & 41 & 0.6101\\
4 & RW \cite{luong2013better} & 1182 & 874 & 0.2714 & 959 & 0.1777 & 753 & 0.2678 & 469 & 0.2161\\
5 & WS-ALL \cite{Finkelstein:2001} & 349 & 142 & 0.6567 & 116 & 0.4071 & 136 & 0.6348 & 74 & 0.5945\\
6 & RG \cite{radovanovic2010hubs} & 62 & 35 & 0.7922 & 14 & 0.3188 & 36 & 0.6411 & 25 & 0.5894\\
7 & MC \cite{miller1991contextual} & 28 & 15 & 0.7675 & 9 & 0.3336 & 16 & 0.2727 & 12 & 0.4706\\
8 & MTurk \cite{halawi2012large} & 283 & 155 & 0.6558 & 123 & 0.5132 & 128 & 0.5591 & 52 & 0.5337\\ \hline
- & Average & 414 & 241 & 0.4402 & 217 & 0.2693 & 219 & 0.4391 & 136 & 0.3612\\ \hline
\end{tabular}
}
\end{table}

In order to have a fair comparison, we reported the accuracy of each approach in two ways in  Table \ref{hqv:table3}. The first accuracy is to compare the coverage and performance of each approach over the all questions in the test dataset (Accuracy All). The second accuracy is to compare the methods over only common questions (Accuracy commons).

Each approach tries to answer as much as possible to the 3,218 questions inside the Phrase Analogy dataset in \textit{Accuracy for All} scenario. For top 30.000 frequent phrases, Google Freebase were able to answer more questions, but for top 3,000,000 frequent phrases ConVec was able to answer more questions with higher accuracy. Fine tuning of the vectors does not have impact on the coverage of ConVec this is why the number of found is similar to the base model. This is mainly because we used the Wikipedia ID of a page instead of its surface name. The heuristic version of ConVec has more coverage to answering questions in comparison with the base ConVec model. The accuracy of the heuristic ConVec is somehow similar to the base ConVec for top 300,000 phrases, but it will drop down for top 3,000,000. It seems that this approach is efficient to increase the coverage without significant sacrificing the accuracy, but probably it needs to be more conservative by adding more regulations and restrictions in the process of adding new anchor texts.

Only common questions between each method are used to compare the \textit{Accuracy for Commons} scenario.
The results in the last column of Table \ref{hqv:table3} show that the fine-tuning of vectors does not have a significant impact on the quality of the vectors embedding. The result of ConVec Heuristic for common questions, argue that this heuristic does not have a significant impact on the quality of base ConVec model and it just improved the coverage (added more concepts to the list of concept vectors). The most important message of the third column of Table \ref{hqv:table3} is that even very small dataset (Wikipedia 2.1 B tokens) is able to produce good vectors embedding in comparison with the Google freebase dataset (100B tokens) and consequently, the quality of the training corpus is more important than its size.

\paragraph{Phrase Similarity Task:}
The next experiment is evaluating vector quality in the Phrase similarity datasets (Check Table \ref{hqv:table4}). In these datasets, each row consists of two words with their relatedness assigned by the human. The Spearman's correlation is used for comparing the result of different approaches with the human evaluated results. These datasets contain words and not the Wikipedia concepts. We replaced all the words in these datasets with their corresponding Wikipedia pages if their surface form and the Wikipedia concept match. We used the simple but effective most frequent sense disambiguation method to disambiguate words that may correspond to several Wikipedia concept. This method of assigning words to concepts is not error prone but this error is considered for all approaches.

\begin{table*}[t]
\centering
\caption{Comparing the results in Phrase Similarity dataset for common entries between all approaches. Rho is Spearmans's correlation.}
\label{hqv:table5}
\resizebox{1\textwidth}{!}{%
\begin{tabular}{lll|c|l|c|c|c}
\hline
\multicolumn{3}{c|}{Datasets} & \multicolumn{1}{l|}{\begin{tabular}[c]{@{}l@{}}Wikipedia \\ Miner\end{tabular}} & \begin{tabular}[c]{@{}l@{}}HitSim\end{tabular} & \multicolumn{1}{l|}{\begin{tabular}[c]{@{}l@{}}ConVec\end{tabular}} & \multicolumn{1}{l|}{\begin{tabular}[c]{@{}l@{}}ConVec\\ (Heuristic)\end{tabular}} & \begin{tabular}[c]{@{}l@{}}ConVec\\ (Only \\ Anchors)\end{tabular} \\ \hline
\# & Dataset Name & \begin{tabular}[c]{@{}l@{}} \#Pairs\end{tabular} & Rho & \multicolumn{1}{c|}{Rho} & Rho & Rho & Rho \\ \hline
1 & WS-REL & 130 & 0.6662 & 0.5330 & 0.6022 & 0.6193 & 0.6515 \\
2 & SIMLEX & 406 & 0.2405 & 0.3221 & 0.3011 & 0.3087 & 0.2503 \\
3 & WS-MAN \cite{Finkelstein:2001}& 224 & 0.6762 & 0.6854 & 0.6331 & 0.6371 & 0.6554 \\
4 & WS-411 \cite{Finkelstein:2001}& 314 & 0.7311 & 0.7131 & 0.7126 & 0.7136 & 0.7308 \\
5 & WS-SIM & 108 & 0.7538 & 0.6968 & 0.7492 & 0.7527 & 0.7596 \\
6 & RWD & 268 & 0.3072 & 0.2906 & 0.1989 & 0.1864 & 0.1443 \\
7 & WS-ALL & 192 & 0.6656 & 0.6290 & 0.6372 & 0.6482 & 0.6733 \\
8 & RG & 20 & 0.7654 & 0.7805 & 0.6647 & 0.7338 & 0.6301 \\
9 & MC & 9 & 0.3667 & 0.5667 & 0.2667 & 0.2167 & 0.2833 \\
10 & MTurk & 122 & 0.6627 & 0.5175 & 0.6438 & 0.6453 & 0.6432 \\ \hline
- & Average & 179 & 0.5333 & 0.5216 & 0.5114 & 0.5152 & 0.5054 \\ \hline
\end{tabular}
}
\end{table*}

Wikipedia Miner \cite{Milne2008} is a well-known approach to find the similarity between two Wikipedia pages based on their input and output links. Results show that our approach for learning concepts embedding can embed the Wikipedia link structure properly since its results is similar to the structural based similarity approach of Wikipedia Miner (See Table \ref{hqv:table4}). The average correlation for the heuristic based approach is less than the other approaches, but average of not-found entries in this approach is much less than the others. It shows that using the heuristic can increase the coverage of the Wikipedia concepts. 

To have a fair comparison between different approaches, we extracted all common entries of all datasets and then re-calculated the correlation (Table \ref{hqv:table5}). We also compared the results with another structural based similarity approach called HitSim \cite{sajadi2015domain}. The comparable result of our approach to structural based methods is another proof that we could embed the Wikipedia link structure properly. The result of heuristic based approach is slightly better than our base model. This shows that without sacrificing the accuracy, we could increase the coverage. This means that with the proposed heuristic, we have a vector representation of more Wikipedia pages.   

Results for only anchors version of ConVec (see the last column of Table \ref{hqv:table5}) show that in some datasets this approach is better than other approaches, but the average result is less than the other approaches. This shows it is better to learn Wikipedia's concepts vector in the context of other words (words that are not anchored) and as a result to have the same vector space for both Concepts and words. 

\section{Conclusion}
In this paper, several approaches for embedding Wikipedia Concepts are introduced. We demonstrated the higher importance of the quality of the corpus than its quantity (size) and argued the idea of the larger corpus will not always lead to a better word embedding. Although the proposed approaches only use inter Wikipedia links (anchors), they have a performance as good as or even higher than the state of the arts approaches for Concept Analogy and Concept Similarity tasks. In contrary to word embedding, Wikipedia Concepts Embedding is not ambiguous, so there is a different vector for concepts with similar surface form but different mentions. This feature is important for many NLP tasks such as Named Entity Recognition, Text Similarity, and Document Clustering or Classification. In the future, we plan to use multiple resources such as Infoboxes, Multilingual Version of a Wikipedia Page, Categories and syntactical features of a page to improve the quality of Wikipedia Concepts Embedding. 

% \bibliographystyle{abbrv}
% \bibliography{ehsansherkat}

\section*{Appendix A: Python Libraries}

\paragraph{} The following libraries are used to extract and prepare the Wikipedia corpus:

\begin{itemize}
\item Wikiextractor: \url{www.github.com/attardi/wikiextractor}
\item Mwparserfromhell: \url{www.github.com/earwig/mwparserfromhell}
\item Wikipedia 1.4.0: \url{www.pypi.python.org/pypi/wikipedia}
\end{itemize}

\paragraph{} The following libraries are used for Word2Vec and Doc2Vec implementation and evaluation: 
\begin{itemize}
\item Gensim: \url{www.pypi.python.org/pypi/gensim}
\item Eval-word-vectors \cite{faruqui-2014}: \url{www.github.com/mfaruqui/eval-word-vectors}
\end{itemize}

\section*{Appendix B: Pruning Wikipedia Pages}
List of rules that are used to prune useless pages from Wikipedia corpus:

\begin{itemize}
\item Having \textless ns0:redirect\textgreater tag in their XML file.
\item There is 'Category:' in the first part of age name.
\item There is 'File:' in the first part of page name.
\item There is 'Template:' in the first part of page name.
\item Anchors having '(disambiguation)' in their page name. Anchors having 'may refer to:' or 'may also refer to' in their text file.
\item There is 'Portal:' in the first part of page name.
\item There is 'Draft:' in the first part of page name.
\item There is 'MediaWiki:' in the first part of page name.
\item There is 'List of' in the first part of the page name.
\item There is 'Wikipedia: in the first part of page name.
\item There is 'TimedText:' in the first part of page name.
\item There is 'Help:' in the first part of page name.
\item There is 'Book:' in the first part of page name.
\item There is 'Module:' in the first part of page name.
\item There is 'Topic:' in the first part of page name.
\end{itemize}

\end{document}